\RequirePackage{amsthm}
\documentclass[sn-mathphys-num]{sn-jnl} 
\usepackage{graphicx}%
\usepackage{multirow}%
\usepackage{amsmath,amssymb,amsfonts}%
\usepackage{mathrsfs}%
\usepackage[title]{appendix}%
\usepackage{xcolor}%
\usepackage{textcomp}%
\usepackage{manyfoot}%
\usepackage{booktabs}%
\usepackage{algorithm}%
\usepackage{algorithmicx}%
\usepackage{algpseudocode}%
\usepackage{listings}%

\usepackage{bm}
\usepackage{csvsimple}
\usepackage{filecontents} 
\usepackage{subcaption}
\usepackage{tikz}
\usepackage{pgfplots}
\usetikzlibrary{positioning,fit}

\pgfplotsset{compat=1.18}
\usetikzlibrary{external}
\usepgfplotslibrary{external}

\newcommand{\candidateset}{\mathcal{S}}

\newcommand{\nrcandidates}{S}
\newcommand{\varnoise}{\sigma}

\newcommand{\newcluster}[1]{\mathcal{C}_{#1}}

\newcommand{\vw}[0]{{\bf w}}
\newcommand{\vx}[0]{{\bf x}}
\newcommand{\vy}[0]{{\bf y}}
\newcommand{\mX}[0]{{\bf X}}

\newcommand\truelabel{y}
\newcommand\labelvec{\vy}
\newcommand\featurevec{\vx}
\newcommand\dataset{\mathcal{D}}
\newcommand{\nodeidx}{i}
\newcommand{\nrnodes}{n}
\newcommand{\weights}{\vw}
\newcommand\hypothesis{h} 
\newcommand{\reward}{r}
\newcommand{\hypospace}{\mathcal{H}}
\newcommand\localmodel[1]{\hypospace^{(#1)}} 
\newcommand{\locallossfunc}[2]{L_{#1}\left(#2 \right)}
\newcommand\learntlocalhypothesis[1]{\widehat{\hypothesis}^{(#1)}}
\newcommand{\localdataset}[1]{\mathcal{D}^{(#1)}}
\newcommand{\localsamplesize}[1]{m_{#1}}
\newcommand{\normgeneric}[2]{\left\Vert  {#1} \right\Vert_{#2}}
\newcommand{\testset}{\dataset^{(\rm t)} }
\newcommand{\lrate}{\eta}
\newcommand{\dimlocalmodel}{d}
\newcommand\defeq{:=}
\newcommand{\nriter}{R}
\newcommand{\itercntr}{k}
\DeclareMathOperator*{\argmin}{argmin}
\newcommand{\pair}[2]{\left( #1,#2 \right)}
\newcommand{\batch}{\mathcal{B}}
\newcommand{\localsampleidx}{r}
\newcommand{\estlocalparams}[1]{\widehat{\mathbf{w}}^{(#1)}}
\newcommand{\samplesize}{m}
\newcommand{\featuremtx}{\mX}
\newcommand{\nodes}{\mathcal{V}}
\newcommand{\clusteridx}{c} 
\newcommand{\nrcluster}{k} 
\newcommand{\cluster}{\mathcal{C}}
\newcommand{\sampleidx}{r}
\newcommand{\learnthypothesis}{\hat{\hypothesis}}

\theoremstyle{thmstyleone}%
\theoremstyle{thmstyletwo}%

\theoremstyle{thmstylethree}%

\newif\ifshowtikz
\showtikztrue

\begin{document}

\title[Article Title]{Your Data, My Model: Learning Who Really Helps in Federated Learning}

%%=============================================================%%
%% GivenName	-> \fnm{Joergen W.}
%% Particle	-> \spfx{van der} -> surname prefix
%% FamilyName	-> \sur{Ploeg}
%% Suffix	-> \sfx{IV}
%% \author*[1,2]{\fnm{Joergen W.} \spfx{van der} \sur{Ploeg} 
%%  \sfx{IV}}\email{iauthor@gmail.com}
%%=============================================================%%

\author[1]{\fnm{Shamsiiat} \sur{Abdurakhmanova}}\email{shamsiiat.abdurakhmanova@aalto.fi}

\author[2]{\fnm{Amirhossein} \sur{Mohammadi}}\email{amirmhd@yorku.ca}

\author[1]{\fnm{Yasmin} \sur{SarcheshmehPour}}\email{yasmin.sarcheshmehpour@aalto.fi}

\author*[1]{\fnm{Alex} \sur{Jung}}\email{alex.jung@aalto.fi}

\affil*[1]{\orgdiv{Department of Computer Science}, \orgname{Aalto University}, \orgaddress{\street{Konemiehentie 2}, \city{Espoo}, \postcode{02150}, \country{Finland}}}

\affil[2]{\orgdiv{Department of Electrical Engineering and Computer Science}, \orgname{York University}, \orgaddress{\street{4700 Keele Street}, \city{Toronto}, \country{Canada}}}

%%==================================%%
%% Sample for unstructured abstract %%
%%==================================%%

\abstract{
Many important machine learning applications involve networks of devices—such as wearables or 
smartphones—that generate local data and train personalized models. A key challenge is determining which 
peers are most beneficial for collaboration. We propose a simple and privacy-preserving method to 
select relevant collaborators by evaluating how much a model improves after a single gradient 
step using another device's data—without sharing raw data. This method naturally extends to 
non-parametric models by replacing the gradient step with a non-parametric generalization. 
Our approach enables model-agnostic, data-driven peer selection for personalized federated learning (PersFL).}


\maketitle

\section{Introduction}\label{Introduction}

Modern federated learning (FL) systems often consist of large, heterogeneous 
networks of data generators, such as users with smartphones or wearable 
devices \cite{HumanActivRegFL}, or abstract entities like diseases in a co-morbidity 
network \cite{NetMedNat2010}. In such systems, personalized federated learning 
(PersFL) aims to train a separate model for each device or data generator, tailored 
to its specific local data and task \cite{PFLMixtures}.

%Consider a network of data generators, such as humans with wearables or smartphones \cite{HumanActivRegFL}. 
%Another, more abstract, example for data generators is diseases connected through a 
%co-morbidity network \cite{NetMedNat2010}. Personalized federated learning (PersFL) 
%aims to train a separate personalized model for each data generator \cite{PFLMixtures}. 

A central challenge in PersFL is data scarcity: local datasets may be too small or too noisy to 
support stand-alone training of complex models \cite{MLBasics}. To address this, devices can 
collaborate with others facing similar tasks. However, collaboration in FL comes with two 
major constraints: privacy (local data cannot be shared freely) and decentralization (data generators 
operate in mobile or distributed environments).

%A key challenge for PersFL is that the available datasets may be too small to independently 
%apply basic machine learning techniques \cite{MLBasics}. In such cases, PersFL can leverage 
%similarities between personal learning tasks to train local models. Unlike basic clustering 
%methods, PersFL methods are typically implemented in distributed (or mobile) computing 
%environments. 

Additionally, PersFL applications often have strict constraints on how much information 
can be shared. In particular, PersFL methods cannot freely access local datasets at 
different locations; instead, they gather information via queries. These queries could 
amount to gradient steps for parametrized models or to more general model updates.  

Unlike traditional clustering methods that rely on Euclidean geometry in feature space 
$\mathbb{R}^m$ \cite[Ch. 8]{MLBasics}, PersFL involves clustering entire learning tasks—
whose similarity is often more abstract and task-specific than simple geometric distance. 
Furthermore, FL applications frequently preclude access to the full network structure 
or a predefined graph of relationships.

%Beyond computational and privacy constraints, a key distinction between basic clustering 
%methods and PersFL is the data geometry. Basic clustering methods, such as $k$-means or 
%Gaussian mixture models, operate on data represented as vectors in Euclidean 
%space $\mathbb{R}^{m}$ \cite[Ch. 8]{MLBasics}. 

%These methods work well when the cluster structure is reflected by the Euclidean distances 
%between numerical feature vectors. By contrast, PersFL involves data points that represent 
%entire learning tasks. The cluster geometry of these learning tasks might deviate significantly 
%from that of ellipsoids in the Euclidean space $\mathbb{R}^{m}$.

\subsection{Our Contribution}
We propose a new PersFL method for adaptively identifying useful collaborators among 
devices that participate in a FL system. Rather than assuming access to a global similarity 
graph or requiring full participation from all devices, our method randomly probes 
other devices (or their data) for their usefulness. For parametric models, 
we implement this probing via monitoring the effect of a gradient step. 
For non-parametric models, we generalize the approach via a proximal step—
corresponding to regularized empirical risk minimization. Our numerical experiments 
indicate that our approach is able to train a high-dimensional personalized model 
by leveraging statistical similarities among small local datasets generated by large 
networks of devices. Section~\ref{sec_persFL_method} introduces the method formally (Algorithm~\ref{alg_pfl_regretmin} and Algorithm~\ref{alg_pfl_regretmin_modelagnostic}). Section~\ref{sec_num_exp} demonstrates 
its effectiveness through numerical experiments.

%{\bf Contribution.} We propose a novel PersFL method that adaptively identifies useful 
%data generators for improving a personalized model. Our method probes randomly selected 
%data generators by analyzing their effect on the incremental training of a personalized model. 
%We then keep only those updates that significantly improve the model performance. 
%Section \ref{sec_persFL_method} spells out this idea as Algorithm \ref{alg_pfl_regretmin} 
%for parametric personalized models. For parametric models, the incremental training amounts 
%to computing gradient steps. We then generalize this method by replacing the gradient step 
%with a proximal step, a form of regularized re-training. The resulting model-agnostic PersFL 
%method for (almost) arbitrary collections of local models is summarized in Algorithm \ref{alg_pfl_regretmin_modelagnostic}. 

\subsection{Related Work} 

\textbf{Graph-Based FL.} A recent line of work uses graph-based methods to exploit similarities 
between local datasets generated by devices of a FL system \cite{Smith2017,GSPBox,JuEusipco2023}. 
Given a similarity (or collaboration) graph, total variation minimization can be used to 
enforce similar model parameters across connected devices with similar data distributions \cite{NetworkLasso}. 
In contrast, our approach does not require on a predefined similarity graph. 
Instead, we infer task similarities in a fully data-driven fashion. In contrast to general graph 
learning methods \cite{pmlr-v202-ye23b}, which attempt to infer the global graph 
structure, our method adopts a local perspective: For a given device, we want to identify 
other devices that could serve as collaborators.

%{\bf Graph-Based FL.} Recent studies in FL uses graphs to represent similarities 
%	between the statistics of local datasets \cite{Smith2017,GSPBox,JuEusipco2023}. 
%	These empirical or collaboration graphs lend naturally to total variation minimization 
%	methods to learn similar local model parameters for similar local datasets \cite{NetworkLasso}. In contrast, our method does not require a predefined graph reflecting similarities. 
%	Rather, we learn similarities between local datasets in a fully data-driven fashion. 
%	Unlike graph learning methods \cite{pmlr-v202-ye23b}, we only learn similarities between a 
%	given local dataset and randomly selected other local datasets. 

\textbf{Clustered FL.} Our method is closely related to clustered FL, which assumes that 
devices form a small number of large clusters \cite{werner2023provably,GhoshCFLTIT,SattlerClusteredFL2020,ClusteredFLTVMinTSP}. 
Devices in the same cluster generate data which can be well-approximated as i.i.d.\ samples 
from a cluster-specific probability distribution. Unlike most clustered FL methods, which 
require global coordination to assign clients to clusters, our method only needs to probe a 
small, randomly selected subset of devices to train a personalized model for a given target 
device.
	
%	{\bf Clustered FL.} Our PersFL method is closely related to clustered FL. Clustered FL 
%	postulates that local datasets, and corresponding learning tasks, form few large clusters \cite{werner2023provably,GhoshCFLTIT,SattlerClusteredFL2020,ClusteredFLTVMinTSP}. 
%	The local datasets in the same cluster can be approximated as i.i.d.\ realizations of a probability distribution. In contrast to the typical clustered FL setting, we do not require to gather information about all data generators. Instead, to train a tailored model for a specific data generator our method randomly picks few other data generators and ranks them according to their effect on the learning progress. 

\textbf{Similarity Measures.} A key difference between our method and clustered FL 
approaches like \cite{werner2023provably,GhoshCFLTIT} lies in how similarity is 
defined and used. Existing methods often rely on Euclidean distances between 
model parameters or their updates. Instead, we define similarity operationally—
as the observed improvement in validation loss when incorporating data from 
another client. This operational notion of similarity is especially useful in FL 
settings involving heterogeneous data and personalized models. As a case in point, 
it is unclear how to compute the distance between model parameters of two personalized 
models using a different number of parameters. This is even more unclear if one of those 
models is non-parametric, such as a random forest. 

\textbf{Probabilistic Models.} We can frame PersFL also as a specific type of probabilistic 
model \cite{Zhu_2023_CVPR}. In particular, we can use a hierarchical probabilistic model 
where each device is associated with a latent variable indicating cluster membership \cite{WuBayesianCFL,2023arXiv230504979K}. 
While we also view datasets as samples from underlying random processes (see Section~\ref{sec_problem_setting}), 
our setting assumes that cluster membership is fixed but unknown—not a latent 
random variable. 

%??? We find that our method requires less tuning compared to 
%direct application of probabilistic inference methods ???? 
%This assumption simplifies model selection and reduces the need 
%for inference over latent structures.
	
%	{\bf Similarity Measures.} A key distinction between our method and the clustered FL methods in 
%	\cite{werner2023provably,GhoshCFLTIT} is in how we measure similarity between local datasets (and their 
%	corresponding learning task). The methods in \cite{werner2023provably,GhoshCFLTIT} use Euclidean distances between local model parameters (or their updates) to cluster similar local datasets. 
%	In contrast, we measure similarity via the effect (on the validation error) of merging local datasets for model training. This construction is appealing for FL applications that use heterogeneous collections data and personalized models. Our method can be applied to collections of parametric and non-parametric models. 

\textbf{Active Sampling and Transfer Learning.} Our method is conceptually similar  
to regret-based active sampling \cite{Baykal2021}, which selects unlabeled data points 
that would yield the greatest benefit if labeled. Instead of distinguishing between labeled 
and unlabeled data, we assume that loss functions corresponding to local tasks form clusters. 
This enables us to identify other tasks whose data can accelerate training for the target 
model—a form of task-level transfer learning \cite{Pan2010}.
	
	%{\bf Active Sampling and Transfer Learning.} Our PersFL method is similar in spirit to a recently proposed regret minimization method for active sampling %\cite{Baykal2021}. This method identifies unlabeled data points that, once labeled, would maximally contribute to the learning progress. We do not %distinguish between labeled and unlabeled data but, rather, assume that the loss functions associated with individual learning tasks form clusters. 
	%Thus, instead of identifying useful data points to train a personalized model we aim at identifying similar learning tasks that provide opportunity for %knowledge transfer \cite{Pan2010}. 
	
\subsection{Outline}
	
Section~\ref{sec_problem_setting} introduces a probabilistic model for local datasets 
generated by networked devices, which we use as a conceptual test-bed for our method. 
Section~\ref{sec_persFL_method} describes our personalized federated learning strategy 
in detail. The key idea is to train a personalized model for a device using selectively chosen 
peer datasets, whose utility is assessed by the reduction in validation error on the target dataset. 
For parametric models, utility is measured using gradient steps; for non-parametric models, 
we use updates via regularized empirical risk minimization. Finally, Section~\ref{sec_num_exp} 
presents numerical experiments demonstrating that our method effectively trains a heterogeneous 
collection of personalized models.

\section{Problem Setting} \label{sec_problem_setting}
	
We consider a collection of $\nrnodes$ devices, indexed as $\nodeidx=1,\ldots,\nrnodes$, 
Each device $\nodeidx$ uses a local dataset $\localdataset{\nodeidx}$ to learn a 
hypothesis $\learntlocalhypothesis{\nodeidx}  \in \localmodel{\nodeidx}$ from a 
local model $\localmodel{\nodeidx}$. The local models can be different for different 
devices, e.g., one device might train a linear model while another device trains a random forest. 
In what follows we focus on the model training at node $\nodeidx=1$ and therefore use 
the shorthands $\widehat{\hypothesis}$ and $\hypospace$ for its learnt hypothesis and 
local model, respectively. 

Our main interest is the high-dimensional regime, where the (effective) dimension 
of the personalized model $\hypospace$ significantly exceeds the size of $\localdataset{1}$. 
This setting is typical when training deep neural networks with many parameters or 
generalized linear models with a large number of features \cite{MLBasics}. To mitigate 
overfitting in such regimes, we regularize the training of $\widehat{\hypothesis}$.

We do so through implicit pooling of other devices’ datasets $\localdataset{\nodeidx}$ that 
share statistical similarity with $\localdataset{1}$. However, privacy constraints in FL 
prevent direct data sharing. Hence, we must assess the utility of $\localdataset{\nodeidx}$ 
for training $\widehat{\hypothesis} \in \hypospace$ via privacy-preserving access mechanisms 
(see Figure \ref{fig_priv_preserving_update}).
	
%We implement this regularization through implicit pooling of local datasets $\localdataset{\nodeidx}$  
%that share similar statistical properties as $\localdataset{1}$. Privacy requirements in FL applications 
%prohibit a naive pooling, such as sharing raw local dataset $\localdataset{\nodeidx}$ 
%to augment the local dataset $\localdataset{1}$. Instead, we must assess the utility of 
%$\localdataset{\nodeidx}$ for learning  $\widehat{\hypothesis} \in \hypospace$ indirectly 
%using privacy-preserving data access mechanisms.

This paper investigates two such mechanisms that enable the use of $\localdataset{\nodeidx}$ 
without sharing raw data. The first mechanism (Section~\ref{sec_parametric_persFL}) applies 
to parametric models such as generalized linear models \cite{MLBasics, JungNetExp2020} and 
deep neural networks \cite{Goodfellow-et-al-2016}. It evaluates the relevance of $\localdataset{\nodeidx}$ 
by computing a gradient step with respect to the current model parameters $\widehat{\weights}$.
	
%This paper studies two different data access mechanisms that allow to leverage information in 
%a local dataset $\localdataset{\nodeidx'}$ without sharing raw data. The first mechanism, studied 
%in Section \ref{sec_parametric_persFL}, applies to parametrized models $\hypospace$, including 
%generalized linear models \cite{MLBasics,JungNetExp2020} or deep neural networks \cite{Goodfellow-et-al-2016}. 
%Roughly speaking, this access mechanism extracts information from a local dataset $\localdataset{\nodeidx}$ by 
%using it to compute a gradient step at the current parameters $\widehat{\weights}$ of the 
%personalized model. 

The second mechanism (Section~\ref{sec_model_agnostic_persfl}) is more general and 
applies to non-parametric model classes. It quantifies the contribution of $\localdataset{\nodeidx}$ 
through a regularized re-training procedure using $\widehat{\weights}$ as initialization, 
again without accessing raw data.

\begin{figure}[h]
	\centering
	\begin{tikzpicture}[
		node/.style={draw, rounded corners, minimum width=2.4cm, minimum height=1.2cm, align=center},
		dataset/.style={draw, circle, minimum size=1cm},
		access/.style={draw, dashed, thick, ->},
		flow/.style={->, thick},
		priv/.style={font=\scriptsize, fill=white},
		highlight/.style={draw=red, very thick}
		]
		
		% Device 1 (central, highlighted)
		\node[dataset] (D1) at (0, 0) {$\localdataset{1}$};
		\node[node, below=0.8cm of D1] (M1) {$\widehat{\hypothesis} \in \hypospace$};
		\node[node, fit={(D1)(M1)}, highlight, inner sep=5pt, label=above:{\textbf{device $\nodeidx=1$}}] (Dev1) {};
		
		\draw[flow] (M1) -- node[right, font=\scriptsize] {val} (D1);
		
		% Device 2
		\node[dataset, left=3.5cm of D1] (D2) {$\localdataset{2}$};
		\node[node, below=0.8cm of D2] (M2) {linear model};
		\node[draw, fit={(D2)(M2)}, inner sep=5pt, label=above:{device $2$}] (Dev2) {};
		
		\draw[flow] (D2) -- (M2);
		\draw[access] (M2.east) to[bend left=15] node[priv] {Sec.\ \ref{sec_parametric_persFL}} (M1.west);
		
		% Device 3
		\node[dataset, right=3.5cm of D1] (D3) {$\localdataset{3}$};
		\node[node, below=0.8cm of D3] (M3) {random forest};
		\node[draw, fit={(D3)(M3)}, inner sep=5pt, label=above:{device $3$}] (Dev3) {};
		
		\draw[flow] (D3) -- (M3);
		\draw[access] (M3.west) to[bend left=15] node[priv] {Sec.\ \ref{sec_model_agnostic_persfl}} (M1.east);
		
		% Padlock icons (or annotate privacy)
	%	\node[font=\scriptsize] at (-1.8, -1.2) { No raw data sharing};
	%	\node[font=\scriptsize] at (1.8, -1.2) {No raw data sharing};
		
	\end{tikzpicture}
	\vspace*{3mm}
	\caption{Personalized model training at device $\nodeidx=1$ via privacy-preserving access to 
		other devices' datasets.\label{fig_priv_preserving_update}}
\end{figure}

%Section \ref{sec_model_agnostic_persfl} discusses a more general access 
%mechanism applicable to non-parametric model $\hypospace$. This access mechanism reveals 
%information about $\localdataset{\nodeidx}$ by using it in regularized re-training of the personalized 
%model $\hypospace$ with the current model parameters $\widehat{\weights}$. 
		
\section{Methods} \label{sec_persFL_method}

Our idea is to learn a personalized hypothesis $\widehat{\hypothesis} \in \hypospace$ 
for device $\nodeidx=1$, using its local dataset $\localdataset{1}$ not as a training set 
but rather as a validation set. This validation set is used to assess the usefulness of other 
nodes' local datasets $\localdataset{\nodeidx'}$, for $\nodeidx' \in \{2,\ldots,\nrnodes\}$. 
Specifically, we simulate an update to the current hypothesis $\widehat{\hypothesis}$ 
using $\localdataset{\nodeidx'}$ and assess the quality of this update via its impact on 
the validation loss.
	
%The main idea of our approach is to learn a tailored hypothesis $\widehat{\hypothesis} \in \hypospace$ 
%for data generator $\nodeidx=1$, using its local dataset $\localdataset{1}$ as a validation set. This 
%validation set is used to score the usefulness of other local datasets $\localdataset{\nodeidx'}$, for 
%some $\nodeidx' \in \{2,\ldots,\nrnodes\}$. In particular, we use $\localdataset{\nodeidx'}$ to update 
%$\widehat{\hypothesis}$ and evaluate the usefulness of this update via the average loss incurred on $\localdataset{1}$. 

%To develop a practical method for personalized federated learning (PersFL), we need to 
%make two aspects precise: 
%(i) how a hypothesis is updated based on an external dataset, and 
%(ii) how the measured utility of such updates informs the learning of the personalized hypothesis $\widehat{\hypothesis}$. 

To operationalise the above idea, we must clarify two aspects: First, we need to choose a 
useful means for updating a hypothesis based on a local dataset. Second, we need to specify 
how the measured similarities between $\localdataset{1}$ and $\localdataset{\nodeidx'}$, for 
$\nodeidx' \in \{2,\ldots,\nrnodes\}$, are eventually exploited for the learning of $\widehat{\hypothesis}$. 
We address these two aspects for two settings:
\begin{itemize}
	\item {\bf Parametric.} Here, we assume that the personalized model $\hypospace$ is parametrized by 
	a vector $\weights \in \mathbb{R}^{\dimlocalmodel}$ (see Section~\ref{sec_parametric_persFL}).
	\item {\bf Model-Agnostic.} Here, we allow for an arbitrary (possibly non-parametric) personalized model $\hypospace$ (see Section~\ref{sec_model_agnostic_persfl}).
\end{itemize}

%We address these questions for two broad regimes 
%that are distinguished by the type of personalized model $\hypospace$. 	
%Section \ref{sec_parametric_persFL} presents a PersFL technique to train a personalized models 
%$\hypospace$ parametrized by a vector $\weights \in \mathbb{R}^{\dimlocalmodel}$. 
%For parametrized models, we can use gradient steps to update $\weights$ using information from 
%a local dataset. Section \ref{sec_model_agnostic_persfl} replaces the gradient step with a more 
%general regularized model (re-) training step. This generalization results in a model-agnostic 
%PersFL technique that also covers non-parametric models such as decision trees \cite{MLBasics,decisiontrees}. 
	
\subsection{Parametric PersFL} 
\label{sec_parametric_persFL}
	
Consider a personalized model $\hypospace$ parametrized by a vector $\weights \in \mathbb{R}^{\dimlocalmodel}$.  
Given a current choice $\weights = \widehat{\weights}$ (e.g, obtained by pre-training \cite{2021arXiv210807258B}), 
we want to further improve (or fine-tune) $\widehat{\weights}$ using the information provided by 
the local datasets $\localdataset{\nodeidx'}$ of devices $\nodeidx'=2,\ldots,\nrnodes$. 

To find out if the local dataset of a device $\nodeidx'$ provides useful information, we simulate a 
gradient step, 
\begin{align} 
\label{equ_def_gradient_step_candidate}
\widetilde{\vw}^{(\nodeidx')} \defeq \widehat{\weights}- \lrate \nabla \locallossfunc{\nodeidx'}{\widehat{\weights}}. 
\end{align}
The gradient step \eqref{equ_def_gradient_step_candidate} updates the current model parameter 
$\widehat{\vw}$ along the direction $-\nabla \locallossfunc{\nodeidx'}{\widehat{\weights}}$. This 
direction is determined by the geometry (or shape) of the local dataset $\localdataset{\nodeidx'}$. 
The extent of the update is controlled by the learning rate $\lrate$. 
     
We quantify the usefulness of the update \eqref{equ_def_gradient_step_candidate} by the reward    
\begin{equation} 
\label{equ_def_reward_generic}
\reward_{\nodeidx'} \defeq \locallossfunc{1}{\widehat{\weights}} - \locallossfunc{1}{\widetilde{\vw}^{(\nodeidx')}}.
\end{equation} 
The reward \eqref{equ_def_reward_generic} is the decrease of the local loss incurred on dataset $\localdataset{1}$ 
when using model parameters $\widetilde{\vw}$ instead of $\widehat{\vw}$. We then implement the 
gradient step \eqref{equ_def_gradient_step_candidate} for the candidate $\nodeidx_{0} \in \candidateset$ 
resulting in the largest reward $\reward_{\nodeidx_{0}} = \max_{\nodeidx'} \reward_{\nodeidx'}$. Trivially, 
$\nodeidx_{0}$ is also obtained by minimizing $\locallossfunc{1}{\widetilde{\vw}^{(\nodeidx')}}$ 
over $\nodeidx' \in \candidateset$. 

We obtain Algorithm \ref{alg_pfl_regretmin} by repeating the above 
procedure for a given number of iterations. 
\begin{algorithm}
		\caption{PersFL for Parametric Models}
		\label{alg_pfl_regretmin}
		\begin{algorithmic}[1]
			\renewcommand{\algorithmicrequire}{\textbf{Input:}}
			\renewcommand{\algorithmicensure}{\textbf{Output:}}
			
			\Require data generators $\nodeidx=1,\ldots,\nrnodes$, each with local 
			loss function $\locallossfunc{\nodeidx}{\cdot}$; learning rate $\lrate$, 
			number $\nriter$ of iterations, number $\nrcandidates$ of candidates. 
		
			\State \label{equ_persfl_param_init} initialize  $\widehat{\weights} \defeq \mathbf{0}$
			
			\For{$\itercntr=0,1,\ldots,\nriter$}
				\vspace*{2mm}
			\State randomly choose set $\candidateset \defeq \left\{ \nodeidx_{1},\ldots,\nodeidx_{\nrcandidates} \right\} \!\subseteq\!\{2,\ldots,\nrnodes\}$ 
			\vspace*{2mm}
		    \State  \label{equ_def_gd_step} for each $\nodeidx'\!\in\!\candidateset$, compute the update (see \eqref{equ_def_gradient_step_candidate})
		     \begin{equation} 
		     	\nonumber
		     	 \widetilde{\weights}^{(\nodeidx')} \defeq \widehat{\weights}- \lrate \nabla \locallossfunc{\nodeidx'}{\widehat{\weights}}
		     \end{equation}

			\State \label{equ_select_candidate_param} determine candidate $\nodeidx_{0} \in \candidateset$ 
			with smallest loss 
			\vspace*{-2mm}
			\begin{equation}
				\nodeidx_{0}  = \argmin_{\nodeidx' \in \candidateset} \locallossfunc{1}{\widetilde{\weights}^{(\nodeidx')}}
			\end{equation}
			\State update $\widehat{\weights}$ via \eqref{equ_def_gradient_step_candidate} for $\nodeidx' = \nodeidx_{0}$
			\EndFor
			\Ensure model parameters  $\widehat{\weights}$ for $\localdataset{1}$
		\end{algorithmic}
	\end{algorithm}
    
Note that Algorithm \ref{alg_pfl_regretmin} requires a differentiable loss function 
$\locallossfunc{\nodeidx}{\cdot}$ for each client $\nodeidx=1,\ldots,\nrnodes$. 
Algorithm \ref{alg_pfl_regretmin} can be adapted for non-differentiable convex loss 
functions by replacing the gradient in step \eqref{equ_def_gd_step} with a 
sub-gradient \cite{NedicTransAC2009}. An alternative to the trivial initialization $\widehat{\weights}= \mathbf{0}$ in 
step \ref{equ_persfl_param_init} is to learn $\widehat{\weights}$ by minimizing 
$\locallossfunc{1}{\cdot}$ (possibly including a penalty term for regularization). 

{\bf Example: Linear Regression.} It is instructive to study the application of 
Algorithm \ref{alg_pfl_regretmin} to train local linear regression models. The local loss function is then 
$\locallossfunc{\nodeidx}{\weights} \defeq (1/\localsamplesize{\nodeidx}) \normgeneric{\vy^{(\nodeidx)} - \mX^{(\nodeidx)}\weights}{2}^{2}$ 
arising in linear regression (see Section \ref{sec_exp_parametric}). Here, the vector 
$\vy^{(\nodeidx)} = \big( \truelabel^{(\nodeidx,1)},\ldots,\truelabel^{(\nodeidx,\localsamplesize{\nodeidx})}\big)^{T}$ 
contains the labels of the data points in the local dataset $\localdataset{\nodeidx}$. 
The rows of the matrix $\mX^{(\nodeidx)} = \big( \featurevec^{(\nodeidx,1)},\ldots,\featurevec^{(\nodeidx,\localsamplesize{\nodeidx})} \big)^{T}$ 
contain the corresponding feature vectors. A basic calculation reveals $\nabla \locallossfunc{\nodeidx}{\weights} = (-2/\localsamplesize{\nodeidx})
\big( \mX^{(\nodeidx)} \big)^{T} \big( \labelvec^{(\nodeidx)} - \mX^{(\nodeidx)} \weights \big)$ and, 
by inserting into \eqref{equ_def_gradient_step_candidate}, step \eqref{equ_select_candidate_param} 
of Algorithm \ref{alg_pfl_regretmin} becomes

\begin{align} 
		\nodeidx_{0}\!&=\!\argmin_{\nodeidx' \in \candidateset} \normgeneric{\vy^{(1)}\!-\!\mX^{(1)} \widetilde{\weights}^{(\nodeidx')}}{2}^{2}   \\ 
		&  \mbox{ with }  \widetilde{\weights}^{(\nodeidx')}  \defeq \widehat{\weights}\!+\!(2\lrate/\localsamplesize{\nodeidx'})
		\big( \mX^{(\nodeidx')} \big)^{T} \big( \labelvec^{(\nodeidx')}\!-\!\mX^{(\nodeidx')} \widehat{\weights}\big). \nonumber
\end{align} 
	
	\subsection{Model-Agnostic PersFL}
	\label{sec_model_agnostic_persfl} 
	
	We now present our second PersFL method that can train an arbitrary model $\hypospace$. 
	Note that Algorithm \ref{alg_pfl_regretmin} is applicable only for parametrized models with differentiable loss functions. This enables the evaluation of the usefulness of $\localdataset{\nodeidx'}$ for training $\hypospace$ through the effect of a gradient step \eqref{equ_def_gradient_step_candidate}. 
	
	Note that the gradient step \eqref{equ_def_gradient_step_candidate} optimizes locally, confined to a neighbourhood around $\widehat{\weights}$, a linear approximation to the loss function $\locallossfunc{\nodeidx'}{\cdot}$. A natural generalization of \eqref{equ_def_gradient_step_candidate} is to locally optimize the loss function itself, 
    
	\begin{align} 
	\label{equ_approx_gd_step}
	\argmin_{\weights \in \mathbb{R}^{\dimlocalmodel}} \lrate \locallossfunc{\nodeidx'}{\weights} + \normgeneric{\weights-\widehat{\weights}}{2}^2. 
	\end{align}

    The symbol $\lrate$ is intentionally reused for the parameter in \eqref{equ_approx_gd_step}, aligning with its use as the step size in \eqref{equ_def_gradient_step_candidate}. A larger value for $\lrate$ in \eqref{equ_approx_gd_step} results in greater progress towards reducing the local loss function $\locallossfunc{\nodeidx'}{\weights}$.  
	For a non-parametric model $\hypospace$ we typically have some means to evaluate and optimize 
	the local loss $\locallossfunc{\nodeidx'}{\hypothesis}$ directly in terms of the hypothesis (instead  
	of its parameters $\locallossfunc{\nodeidx'}{\weights}$).\footnote{Indeed, without any means of 
	optimizing $\locallossfunc{\nodeidx'}{\hypothesis}$ over $\hypothesis \in \hypospace$, 
	training the model $\hypospace$ would be impossible. Such means of optimizing $\locallossfunc{\nodeidx'}{\hypothesis}$ are provided by software libraries such as the Python package \texttt{scikit-learn} \cite{JMLR:v12:pedregosa11a}.}
	
	It remains to replace the second term in \eqref{equ_approx_gd_step} with some quantitative measure for the deviation between two hypothesis maps $\hypothesis, \widehat{\hypothesis} \in \hypospace$. This measure can be derived by comparing their predictions on an unlabeled test set $\testset$. For numeric labels, the squared difference between predictions of the hypothesis maps can be used to quantify their deviation, i.e., 
    
	\begin{equation}
		\label{equ_measure_test_set_pred_error}
	 (1/|\testset|)	\sum_{\featurevec \in  \testset} \big( \hypothesis(\featurevec) -  \widehat{\hypothesis}(\featurevec) \big)^2.  
	\end{equation} 	
This leads to a generalization of the gradient step \eqref{equ_def_gradient_step_candidate} by 
inserting \eqref{equ_measure_test_set_pred_error} into \eqref{equ_approx_gd_step} and formulating 
the objective directly in terms of the hypothesis $\hypothesis \in \hypospace$

	\begin{align} 
		\label{equ_def_update_modelagnostic}
  \hspace*{-2mm}\widetilde{\hypothesis}^{(\nodeidx')}\!=\!\argmin_{\hypothesis \in \hypospace} \hspace*{-1mm} \bigg[   \lrate \locallossfunc{\nodeidx'}{\hypothesis}\!+\!(1/|\testset|)	\hspace*{-2mm}\sum_{\featurevec \in  \testset} \hspace*{-2mm} \big( \hypothesis(\featurevec)\!-\!\widehat{\hypothesis}(\featurevec) \big)^2 \bigg]. 
	\end{align} 
    
	The objective function in \eqref{equ_def_update_modelagnostic} consists of two components that 
	have counteracting effects on the updated hypothesis $\widetilde{\hypothesis}^{(\nodeidx')}$: 
	\begin{itemize} 
		\item The first component ensures that the hypothesis $\widetilde{\hypothesis}^{(\nodeidx')}$ performs 
	well on the local dataset $\localdataset{\nodeidx'}$ of node $\nodeidx'$.\footnote{Note that we do 
	not require direct access to $\localdataset{\nodeidx'}$ but only via the local loss function $\locallossfunc{\nodeidx'}{\cdot}$.}
	\item The second component requires $\widetilde{\hypothesis}^{(\nodeidx')}$ to deliver similar predictions 
	as the hypothesis $\widehat{\hypothesis}$ on the test set $\testset$.
	\end{itemize}
	
We arrive at Algorithm \ref{alg_pfl_regretmin_modelagnostic} as a generalization of 
Algorithm \ref{alg_pfl_regretmin} by replacing the gradient step \eqref{equ_def_gradient_step_candidate} 
with the update step \eqref{equ_def_update_modelagnostic}. 

\begin{algorithm}
\caption{Model Agnostic PersFL}
\label{alg_pfl_regretmin_modelagnostic}
\begin{algorithmic}[1]
\renewcommand{\algorithmicrequire}{\textbf{Input:}}
\renewcommand{\algorithmicensure}{\textbf{Output:}}
			
\Require data generators with local loss function $\locallossfunc{\nodeidx}{\cdot}$, for $\nodeidx=1,\ldots,\nrnodes$, 
           global test-set $\testset$, learning rate $\lrate$, number $\nriter$ of iterations, candidate set size $\nrcandidates$, 
           personal model $\hypospace$.  
			
\State  initialize hypothesis $\widehat{\hypothesis} \in \hypothesis$ 

\For{$\itercntr=0,1,\ldots,\nriter$}

\vspace*{2mm}
\State randomly choose set $\candidateset \defeq \left\{ \nodeidx_{1},\ldots,\nodeidx_{\nrcandidates} \right\} \!\subseteq\!\{2,\ldots,\nrnodes\}$ 
\vspace*{2mm}
\State  for each $\nodeidx'\!\in\!\candidateset$, compute $\widetilde{\hypothesis}^{(\nodeidx')}$ via \eqref{equ_def_update_modelagnostic}
%\vspace*{2mm}
%\State for each $\nodeidx'\!\in\!\candidateset$, compute loss $\ell_{\nodeidx'} = \locallossfunc{1}{\widetilde{\hypothesis}^{(\nodeidx')}}$
\vspace*{2mm}
\State determine candidate  $\nodeidx_{0} \in \candidateset$ with smallest loss 
\vspace*{-2mm}
\begin{equation}
	\ell_{\nodeidx_{0}}  = \argmin_{\nodeidx' \in \candidateset}  \ell_{\nodeidx'}  
\end{equation}
\State update $\widehat{\hypothesis}$ via \eqref{equ_def_update_modelagnostic} for $\nodeidx' = \nodeidx_{0}$
%\State aggregate $\localtestdataset{\nodeidx_{0}}$ to $\mathcal{D}'^{(1)}$  \label{op8_alg2}
\EndFor
\Ensure learnt hypothesis  $\widehat{\hypothesis}$ for $\localdataset{1}$
\end{algorithmic}
\end{algorithm}	
Note that Algorithm \ref{alg_pfl_regretmin_modelagnostic} requires the specification of a  
loss function $\locallossfunc{\nodeidx}{\cdot}$ for each client $\nodeidx=1,\ldots,\nrnodes$. 
Moreover, we also need to specify how to initialize the local model $\widehat{\hypothesis}$. 
One option is to initialize $\widehat{\hypothesis}$ by plain model training on the local dataset $\localdataset{1}$ (see Section \ref{sec_exp_non_parametric}). 

{\bf Non-parametric Least-Squares.} Consider the application of Algorithm \ref{alg_pfl_regretmin_modelagnostic} 
to non-parametric least-squares regression. Here, a local loss function of the form $\locallossfunc{\nodeidx}{\hypothesis}  = (1/\localsamplesize{\nodeidx}) \sum_{\pair{\featurevec}{\truelabel}} \big(\truelabel - \hypothesis(\featurevec) \big)^2 $ is employed 
for each client $\nodeidx=1,\ldots,\nrnodes$. In this case, the update \eqref{equ_def_update_modelagnostic} 
corresponds to a form of data augmentation \cite[Ch. 7]{MLBasics}. Indeed, \eqref{equ_def_update_modelagnostic} 
is equivalent to the training of $\hypospace$ on the local dataset $\localdataset{\nodeidx'}$ 
augmented with the data points $\pair{\featurevec}{\widehat{\hypothesis}(\featurevec)}$, $\featurevec \in \testset$. 
The data points in the resulting augmented dataset are weighted by $(1/\localsamplesize{\nodeidx})$ 
and $\lrate$, respectively. 

\begin{figure} 
\begin{center}
\ifshowtikz
\begin{tikzpicture}[scale=1]
	\draw[->, very thick] (0,0.5) -- (7.7,0.5) node[above,xshift=-5pt,yshift=5pt] {feature $\featurevec$};       % X-axis
	\draw[->, very thick] (0.5,0) -- (0.5,4.2) node[above] {label $\truelabel$};   % Y-axis
	
	\draw[color=black, thick, dashed, domain = -0.5: 5.2, variable = \x]  plot ({\x},{\x*0.4 + 2.0}) ;     
	\node at (5.7,4.1) {$\hypothesis(\featurevec)$};    
	
	\coordinate (l1)   at (1.2, 2.48);
	\coordinate (l2) at (1.4, 2.56);
	\coordinate (l3)   at (1.7,  2.68);
	
	\coordinate (l4)   at (2.2, 2.2*0.4+2.0);
	\coordinate (l5) at (2.4, 2.4*0.4+2.0);
	\coordinate (l6)   at (2.7,  2.7*0.4+2.0);
	
	\coordinate (l7)   at (3.9,  3.9*0.4+2.0);
	\coordinate (l8) at (4.2, 4.2*0.4+2.0);
	\coordinate (l9)   at (4.5,  4.5*0.4+2.0);
	
	\coordinate (n1)   at (1.2, 1.8);
	\coordinate (n2) at (1.4, 1.8);
	\coordinate (n3)   at (1.7,  1.8);
	
	\coordinate (n4)   at (2.2, 3.8);
	\coordinate (n5) at (2.4, 3.8);
	\coordinate (n6)   at (2.7,  3.8);

	% augemented data point obtained by perturbing feature, not touching label value 
	\coordinate (n7)   at (3.9, 2.6);
	\coordinate (n8) at (4.2, 2.6);
	\coordinate (n9)   at (4.5,  2.6);
	
	\node at (n1)  [circle,draw,fill=red,minimum size=6pt,scale=0.6, name=c1] {};
	\node at (n2)  [circle,draw,fill=blue,minimum size=6pt, scale=0.6, name=c2] {};
	\node at (n3)  [circle,draw,fill=red,minimum size=6pt,scale=0.6,  name=c3] {};
	\node at (n4)  [circle,draw,fill=red,minimum size=12pt, scale=0.6, name=c4] {};  
	\node at (n5)  [circle,draw,fill=blue,minimum size=12pt,scale=0.6,  name=c5] {};
	\node at (n6)  [circle,draw,fill=red,minimum size=12pt, scale=0.6, name=c6] {};  

	\draw[<->, color=red, thick] (l1) -- (c1);  
	\draw[<->, color=blue, thick] (l2) -- (c2);  
	\draw[<->, color=red, thick] (l3) -- (c3);  
	\draw[<->, color=red, thick] (l4) -- (c4);  
	\draw[<->, color=blue, thick] (l5) -- (c5);  
	\draw[<->, color=red, thick] (l6) -- (c6);  
	
	\draw[fill=blue] (5.0, 3.2)  circle (0.1cm) node [black,xshift=0.6cm] {$\localdataset{\nodeidx'}$};
	\draw[fill=red] (5.0, 2.5)  circle (0.1cm) node [black,xshift=1.8cm] {$\pair{\featurevec}{\widehat{\hypothesis}(\featurevec)}$, $\featurevec \in \testset$};
\end{tikzpicture}
   \else
% Placeholder for the TikZ figure
\fbox{\parbox{0.5\textwidth}{TikZ figure suppressed.}}
\fi
\end{center}
\vspace*{3mm}
\caption{Algorithm \ref{alg_pfl_regretmin_modelagnostic} uses a generalized gradient step \eqref{equ_def_update_modelagnostic}
	to update a non-parametric model $\widehat{\hypothesis} \in \hypospace$. This update 
	can be interpreted as a form of regularization via data augmentation. 
	%Indeed, the model re-training step \eqref{equ_def_update_modelagnostic} amounts 
	%to training $\hypospace$ using the local dataset $\localdataset{\nodeidx'}$ augmented by the data points $\pair{\featurevec}{\widehat{\hypothesis}(\featurevec)}$, $\featurevec \in \testset$.
	 \label{equ_def_data_aug_non_par_regression}}
\end{figure} 

    \subsection{Online Variants} 
\label{sec_online_variants} 

Note that both, Algorithm \ref{alg_pfl_regretmin} and Algorithm \ref{alg_pfl_regretmin_modelagnostic}, 
require the complete local loss function for each client $\nodeidx = 1,\ldots,\nrnodes$ as input. In 
some applications, providing these loss functions as input to the algorithms may be difficult or even infeasible. For instance, the loss functions might only be accessible as averages over local datasets $\localdataset{\nodeidx'}$ that are generated sequentially over time. 

Online variants of Algorithm \ref{alg_pfl_regretmin} and Algorithm \ref{alg_pfl_regretmin_modelagnostic} initiate 
a new iteration whenever a sufficient amount of new data points have been generated. 

In each iteration $\itercntr$, an online variant of Algorithm \ref{alg_pfl_regretmin} computes a 
new estimate $\tilde{L}^{(\itercntr)}_{\nodeidx}(\cdot)$ for the loss function $\locallossfunc{\nodeidx'}{\cdot}$. The estimate, $\tilde{L}^{(\itercntr)}_{\nodeidx}(\cdot)$, can be computed as the average loss over the most recent batch $\batch$ of data points.

\section{Numerical Experiments} 
\label{sec_num_exp}

This section presents numerical experiments illustrating the effectiveness of Algorithm \ref{alg_pfl_regretmin} for parametrized models (Section \ref{sec_exp_parametric}) and Algorithm \ref{alg_pfl_regretmin_modelagnostic} 
for non-parametric models (Section \ref{sec_exp_non_parametric}). %Synthetic local datasets are used in both sections.
The code for these experiments is available in the corresponding \href{https://github.com/shamPJ/PersFL/blob/main/persFL.ipynb}{GitHub repository}.

\subsection{Toy Dataset} 
\label{sec_exp_parametric}
To benchmark our method we generate a synthetic dataset as follows: 
Local datasets $\localdataset{\nodeidx}$ consist of $\localsamplesize{\nodeidx}$ 
realizations of i.i.d. random variables with a common probability distribution $p^{(\nodeidx)}$, 
% The local dataset $\localdataset{\nodeidx}$ consists of $\localsamplesize{\nodeidx}$ individual data points, 
\begin{equation}
	\label{equ_def_local_dataset}
	\localdataset{\nodeidx} = \bigg\{ \pair{ \featurevec^{(\nodeidx,1)}}{\truelabel^{(\nodeidx,1)}}, \ldots \pair{ \featurevec^{(\nodeidx,\localsamplesize{\nodeidx})}}{\truelabel^{(\nodeidx,\localsamplesize{\nodeidx})}}\bigg\}.
\end{equation}
Each data point is characterized by a feature vector $\featurevec^{(\nodeidx,\localsampleidx)} \in \mathbb{R}^{\dimlocalmodel}$ 
and a scalar label $\truelabel^{(\nodeidx,\localsampleidx)}$, for $\localsampleidx=1,\ldots,\localsamplesize{\nodeidx}$. 
It is convenient to stack the feature vectors and labels of a local dataset $\localdataset{\nodeidx}$, 
into a feature matrix and label vector, respectively, 
\begin{align}
	\featuremtx^{(\nodeidx)}& \defeq \big( \featurevec^{(\nodeidx,1)}, \ldots,\featurevec^{(\nodeidx,\localsamplesize{\nodeidx})}  \big)^{T} \in \mathbb{R}^{\localsamplesize{\nodeidx} \times \dimlocalmodel}  \mbox{, and } \nonumber \\ 
	\labelvec^{(\nodeidx)} & \defeq \big( \truelabel^{(\nodeidx,1)}, \ldots, \truelabel^{(\nodeidx,\localsamplesize{\nodeidx})}\big)^{T} \in \mathbb{R}^{\localsamplesize{\nodeidx}}.
\end{align} 

The feature vectors $ \featurevec^{(\nodeidx,\sampleidx)} \sim \mathcal{N}(\mathbf{0},\mathbf{I}_{\dimlocalmodel \times \dimlocalmodel})$ are drawn i.i.d.\ from a standard multivariate normal distribution. The labels $\truelabel^{(\nodeidx,\sampleidx)}$ 
of the data points, stacked into the vector $\vy^{(\nodeidx)} \in \mathbb{R}^{\localsamplesize{\nodeidx}}$, are generated via a noisy linear model, 
\begin{equation} 
	\label{equ_def_true_linear_model_SBM}
	\vy^{(\nodeidx)} = \featuremtx^{(\nodeidx)} \overline{\weights}^{(\nodeidx)} + \varnoise  {\bm \varepsilon}^{(\nodeidx)}. 
	\vspace{0mm}
\end{equation} 
The noise terms $ {\bm \varepsilon}^{(\nodeidx)} \sim  \mathcal{N}(\mathbf{0} , \mathbf{I} )$, for $\nodeidx \in \nodes$ 
are drawn i.i.d.\ from a standard normal distribution. The noise variance $\varnoise^{2} \geq 0$ in \eqref{equ_def_true_linear_model_SBM} 
is assumed fixed.

The clients are partitioned into two disjoint clusters 
\begin{equation} 
\label{equ_def_cluster_partition} 
\nodes = \newcluster{1} \cup  \newcluster{2} \mbox{ , with identical cluster sizes } \nrnodes_{1}=\nrnodes_{2}. 
\end{equation} 
The underlying true model parameters $\overline{\weights}^{(\nodeidx)}$ in \eqref{equ_def_true_linear_model_SBM} 
are identical for nodes within the same cluster, i.e., $\overline{\weights}^{(\nodeidx)} = \overline{\weights}^{(1)}$ 
for $\nodeidx  \in \newcluster{1}$ and $\overline{\weights}^{(\nodeidx)} = \overline{\weights}^{(2)}$ 
for $\nodeidx  \in \newcluster{2}$. For each cluster $\clusteridx \in \{1,2\}$, we generate a corresponding 
parameter vector $\weights^{(\clusteridx)}$ by filling its entries with i.i.d. samples from the uniform distribution $\mathcal{U}_{[-5, 5]}$. 

\subsection{Training a Personalized Linear Model} 
We use Algorithm \ref{alg_pfl_regretmin} to learn local model parameters $\weights$ of a linear 
hypothesis $\hypothesis(\featurevec) \defeq \weights^{T} \featurevec$ for the data generator 
$\nodeidx=1$. The loss incurred by a model parameters $\weights$ on dataset $\localdataset{\nodeidx}$ is 

\begin{align}
	\label{equ_def_suqared_error_loss_linmodel}
	\locallossfunc{\nodeidx}{\weights} & \defeq (1/\localsamplesize{\nodeidx}) \sum_{\sampleidx=1}^{\localsamplesize{\nodeidx}} \big( \truelabel^{(\nodeidx,\sampleidx)} - \weights^T \featurevec^{(\nodeidx,\sampleidx)}   \big)^2 \nonumber \\ 
	& = (1/\localsamplesize{\nodeidx}) \normgeneric{\vy^{(\nodeidx)} - \featuremtx^{(\nodeidx)} \weights }{2}^2.
\end{align} 

For the input to Algorithm \ref{alg_pfl_regretmin} we generate $\nrnodes = 100$ local 
datasets according to \eqref{equ_def_true_linear_model_SBM} with $\localsamplesize{\nodeidx}=\localsamplesize{}=10$ 
and varying dimension $\dimlocalmodel$. The local datasets are clustered into two equal-sized 
clusters, $\newcluster{1}$ and $\newcluster{2}$ with $\nrnodes_{1}=\nrnodes_{2}=50$. 

We measure the quality of the learnt local model parameter $\estlocalparams{\nodeidx}$ 
using the squared Euclidean norm 

\begin{equation}
		\label{equ_def_MSE}
{\rm MSE} \defeq 	\normgeneric{\widehat{\weights}-  \overline{\weights}^{(1)}}{2}^{2}.
\end{equation}

Figure \ref{fig:alg_pfl_regretmin_params} shows the MSE \eqref{equ_def_MSE} for the client $\nodeidx=1$ 
obtained from Algorithm \ref{alg_pfl_regretmin} with different parameters and a fixed learning rate $\lrate=0.05$. 

Figure \ref{fig:param_dm} indicates slower convergence as the dimension $\dimlocalmodel$ of the local linear model increases. Figure \ref{fig:param_noise} shows that Algorithm \ref{alg_pfl_regretmin} incurs higher MSE with increasing noise levels $\varnoise$ in the local linear model \eqref{equ_def_true_linear_model_SBM}. Figure \ref{fig:param_subsize} highlights that the size $\nrcandidates$ of the random candidate subset $\candidateset^{(\itercntr)}$ must be sufficiently large to include a few clients from the cluster associated with client $\nodeidx=1$. 

\begin{figure}[htbp]
    \centering
    \begin{subfigure}[b]{0.3\textwidth}
        \centering
        \ifshowtikz
		\begin{tikzpicture}[scale=0.5]
			\begin{axis}[
				legend style={nodes={scale=0.8}},
				xlabel={iteration nr.\ $\itercntr$},
				xlabel style={font=\fontsize{15}{10.8}\selectfont},
				ylabel={},
				ymax=1e4,
				tick label style={font=\fontsize{15}{15}\selectfont}, % Custom tick label size in pt
				ymajorgrids=true,
				grid style=dashed,
				table/col sep=comma,
				ymode=log,
				legend style={
					at={(0.97,0.6)},
					anchor=north east,
					font=\fontsize{15}{9.6}\selectfont,
					nodes={scale=1}
				},
				cycle list={
					cyan!40!blue!60,
					orange,
					black!70!green!70,
					purple,
					violet
				},
				]   
				\foreach \i/\j in {1/0.2, 2/1, 3/2, 4/5, 5/10}{
					\edef\temp{\noexpand\addlegendentry{$\dimlocalmodel/\localsamplesize{\nodeidx}$ = \j}
						\noexpand\addplot+ [mark=none, very thick, solid] table [
						x index=0, 
						y index=\i] {Algo1_dm_ratio.csv};
					}
					\temp}
			\end{axis}
		\end{tikzpicture}
	    \caption{$\varnoise=0$, $\nrcandidates=20$.}
	\label{fig:param_dm}
    \end{subfigure} \hfill
    \begin{subfigure}[b]{0.3\textwidth}
    \centering
    \ifshowtikz
    \begin{tikzpicture}[scale=0.5]
    \begin{axis}[
    	legend style={nodes={scale=0.8}},
    	xlabel={iteration nr.\ $\itercntr$},
    	xlabel style={font=\fontsize{15}{10.8}\selectfont},
    	ylabel={},
    	tick label style={font=\fontsize{15}{15}\selectfont}, % Custom tick label size in pt
    	ymajorgrids=true,
    	grid style=dashed,
    	ymin=0.5e-3,     % <-- lower limit
    	ymax=5e2,      % <-- upper limit
    	table/col sep=comma,
    	ymode=log,
    	legend style={
    		at={(0.97,0.99)},
    		anchor=north east,
    		font=\fontsize{15}{9.6}\selectfont,
    		nodes={scale=1}
    	},
    	cycle list={
    		cyan!40!blue!60,
    		orange,
    		black!70!green!70,
    		purple,
    		violet
    	},
    	]           
				\foreach \i/\j in {1/0.05, 2/0.1, 3/0.2, 4/0.5, 5/1}{
					\edef\temp{\noexpand\addlegendentry{$\varnoise$ = \j}
						\noexpand\addplot+ [mark=none, very thick, solid] table [
						x index=0, 
						y index=\i] {Algo1_noise.csv};
					}
					\temp}
			\end{axis}
    \end{tikzpicture}
     \caption{$\dimlocalmodel/\localsamplesize{}=2$, $\nrcandidates=20$.}
        \label{fig:param_noise}
    \end{subfigure} \hfill
    \begin{subfigure}[b]{0.3\textwidth}
    \centering
    \ifshowtikz
    \begin{tikzpicture}[scale=0.5]
        \begin{axis}[
                legend style={nodes={scale=0.8}},
				xlabel={iteration nr.\ $\itercntr$},
				xlabel style={font=\fontsize{15}{10.8}\selectfont},
				ylabel={},
			   tick label style={font=\fontsize{15}{15}\selectfont}, % Custom tick label size in pt
				ymajorgrids=true,
				grid style=dashed,
				table/col sep=comma,
				ymode=log,
				legend style={
					at={(0.97,0.6)},
					anchor=north east,
					font=\fontsize{15}{9.6}\selectfont,
					nodes={scale=1}
				},
				cycle list={
					cyan!40!blue!60,
					orange,
					black!70!green!70,
					purple,
					violet
				},
				]   
				\foreach \i/\j in {1/5, 2/10, 3/15, 4/20, 5/30}{
					\edef\temp{\noexpand\addlegendentry{$\nrcandidates\!=$\j}
						\noexpand\addplot+ [mark=none, very thick, solid] table [
						x index=0, 
						y index=\i] {Algo1_subset_size.csv};
					}
					\temp}
			\end{axis}
    \end{tikzpicture}
    \caption{$\dimlocalmodel/\localsamplesize{\nodeidx}=2$, $\varnoise=0$.}
        \label{fig:param_subsize}
    \end{subfigure} \hfill
\caption{MSE incurred by Algorithm \ref{alg_pfl_regretmin} for varying model and algorithm 
	hyper-parameters: (a) varying $\dimlocalmodel/\localsamplesize{}$, (b) varying $\varnoise$, 
	and (c) varying $\nrcandidates$. The source code for the experiment can be found at \cite{ShamsiRepo}.}
\label{fig:alg_pfl_regretmin_params}
\end{figure}

\subsection{Comparison with IFCA}

This experiment compares Algorithm \ref{alg_pfl_regretmin} with the recently proposed 
\emph{Iterative Federated Clustering Algorithm} (IFCA) \cite{GhoshCFLTIT}. 
Figure \ref{fig:ifca} illustrates the MSE incurred by both methods on the synthetic linear regression data \eqref{equ_def_true_linear_model_SBM},  with varying $\dimlocalmodel/\samplesize$ ratio. Using the same learning rate $\lrate = 0.05$, Algorithm \ref{alg_pfl_regretmin} performs comparably to IFCA. Overall, IFCA requires fewer iterations when applied to local datasets \eqref{equ_def_true_linear_model_SBM} with increasing ratio $\dimlocalmodel/\samplesize$.

\begin{figure*}
	\centering 
	\begin{subfigure}[b]{0.3\textwidth}
		\ifshowtikz
		\begin{tikzpicture}[scale=0.5]
			\begin{axis}[
				axis lines = left,
				enlarge x limits = true,
				enlarge y limits = true,
				xlabel={iteration nr.\ $\itercntr$},
				ylabel={MSE},
				ymajorgrids=true,
				grid style=dashed,
				table/col sep=comma,
				ymode=log,
				cycle list={
					cyan!40!blue!60,
					orange}]   
				\addplot+ [mark=none, very thick, solid] table [x index=0,
				y index=1] {Algo1vsIfca_Algo1_baseline.csv};
				\addplot+ [mark=none, very thick, solid] table [x index=0,
				y index=1] {Algo1vsIfca_ifca_baseline.csv};
			\end{axis}
		\end{tikzpicture}
		   \else
		% Placeholder for the TikZ figure
		\fbox{\parbox{0.5\textwidth}{TikZ figure suppressed.}}
		\fi
		\caption{$\dimlocalmodel/\samplesize=0.2$}
	\end{subfigure}\hfill
	\begin{subfigure}[b]{0.3\textwidth}
		\ifshowtikz
		\begin{tikzpicture}[scale=0.5]
			\begin{axis}[
				axis lines = left,
				enlarge x limits = true,
				enlarge y limits = true,
				xlabel={iteration nr.\ $\itercntr$},
				ymajorgrids=true,
				grid style=dashed,
				table/col sep=comma,
				ymode=log,
				cycle list={
					cyan!40!blue!60,
					orange}]   
				\addplot+ [mark=none, very thick, solid] table [x index=0,
				y index=2] {Algo1vsIfca_Algo1_baseline.csv};
				\addplot+ [mark=none, very thick, solid] table [x index=0,
				y index=2] {Algo1vsIfca_ifca_baseline.csv};
			\end{axis}
		\end{tikzpicture}
		   \else
		% Placeholder for the TikZ figure
		\fbox{\parbox{0.5\textwidth}{TikZ figure suppressed.}}
		\fi
		\caption{$\dimlocalmodel/\samplesize=2$}
	\end{subfigure}\hfill
	\begin{subfigure}[b]{0.3\textwidth}
		\ifshowtikz
		\begin{tikzpicture}[scale=0.5]
			\begin{axis}[
				axis lines = left,
				enlarge x limits = true,
				enlarge y limits = true,
				xlabel={iteration nr.\ $\itercntr$},
				ymajorgrids=true,
				grid style=dashed,
				table/col sep=comma,
				ymode=log,
				cycle list={
					cyan!40!blue!60,
					orange}]
				\addplot+ [mark=none, very thick, solid] table [x index=0,
				y index=3] {Algo1vsIfca_Algo1_baseline.csv};
				\addlegendentry{Algorithm 1}
				\addplot+ [mark=none, very thick, solid] table [x index=0,
				y index=3] {Algo1vsIfca_ifca_baseline.csv};
				\addlegendentry{IFCA}
			\end{axis}
		\end{tikzpicture}
		   \else
		% Placeholder for the TikZ figure
		\fbox{\parbox{0.5\textwidth}{TikZ figure suppressed.}}
		\fi
		\caption{$\dimlocalmodel/\samplesize=5$}
	\end{subfigure}
	\caption{MSE incurred by Algorithm \ref{alg_pfl_regretmin} and IFCA for local 
		datasets forming $\nrcluster=2$ clusters (see \eqref{equ_def_cluster_partition}). IFCA 
		is provided the correct number of clusters.}
	\label{fig:ifca}
\end{figure*}

A key distinction between Algorithm \ref{alg_pfl_regretmin} and IFCA is that Algorithm \ref{alg_pfl_regretmin} does not require prior knowledge of the number of clusters formed by the data generators (see \eqref{equ_def_cluster_partition}). In contrast, IFCA requires 
the number of clusters as an input parameter. Figure \ref{fig:ifca_no_clusters_not_known} 
compares the performance of Algorithm \ref{alg_pfl_regretmin} with IFCA when using an incorrect number of clusters, i.e., different from the number $\nrcluster=5$ of clusters  in \eqref{equ_def_true_linear_model_SBM}. As shown in Figure \ref{fig:ifca_no_clusters_not_known}, 
Algorithm \ref{alg_pfl_regretmin} significantly outperforms IFCA when the number of clusters is misspecified. 

\begin{figure*}
\captionsetup[subfigure]{justification=centering}
\centering 
\begin{subfigure}[b]{0.3\textwidth}
	\begin{tikzpicture}[scale=0.5]
		\begin{axis}[
			axis lines = left,
			enlarge x limits = true,
			enlarge y limits = true,
			xlabel={iteration nr.\ $\itercntr$},
			ylabel={MSE},
			ymajorgrids=true,
			grid style=dashed,
			table/col sep=comma,
			ymode=log,
			cycle list={
				cyan!40!blue!60,
				orange}]   
			\addplot+ [mark=none, very thick, solid] table [x index=0,
			y index=1] {Algo1vsIfca_Algo1_cl_mismatch.csv};
			\addplot+ [mark=none, very thick, solid] table [x index=0,
			y index=1] {Algo1vsIfca_ifca_cl_mismatch.csv};
		\end{axis}
	\end{tikzpicture}
	\caption{$\dimlocalmodel/\samplesize=0.2$}
\end{subfigure}\hfill
\begin{subfigure}[b]{0.3\textwidth}
	\begin{tikzpicture}[scale=0.5]
		\begin{axis}[
			axis lines = left,
			enlarge x limits = true,
			enlarge y limits = true,
			xlabel={iteration nr.\ $\itercntr$},
			ymajorgrids=true,
			grid style=dashed,
			table/col sep=comma,
			ymode=log,
			cycle list={
				cyan!40!blue!60,
				orange}]   
			\addplot+ [mark=none, very thick, solid] table [x index=0,
			y index=2] {Algo1vsIfca_Algo1_cl_mismatch.csv};
			\addplot+ [mark=none, very thick, solid] table [x index=0,
			y index=2] {Algo1vsIfca_ifca_cl_mismatch.csv};
		\end{axis}
	\end{tikzpicture}
	\caption{$\dimlocalmodel/\samplesize=2$}
\end{subfigure}\hfill
\begin{subfigure}[b]{0.3\textwidth}
	\begin{tikzpicture}[scale=0.5]
		\begin{axis}[
			axis lines = left,
			enlarge x limits = true,
			enlarge y limits = true,
			xlabel={iteration nr.\ $\itercntr$},
			ymajorgrids=true,
			grid style=dashed,
			table/col sep=comma,
			ymode=log,
			cycle list={
				cyan!40!blue!60,
				orange}]
			\addplot+ [mark=none, very thick, solid] table [x index=0,
			y index=3] {Algo1vsIfca_Algo1_cl_mismatch.csv};
			\addlegendentry{Algorithm \ref{alg_pfl_regretmin}}
			\addplot+ [mark=none, very thick, solid] table [x index=0,
			y index=3] {Algo1vsIfca_ifca_cl_mismatch.csv};
			\addlegendentry{IFCA}
		\end{axis}
	\end{tikzpicture}
	\caption{$\dimlocalmodel/\samplesize=5$}
\end{subfigure}
\caption{MSE incurred by Algorithm \ref{alg_pfl_regretmin} and IFCA for local 
	datasets forming $\nrcluster=5$ clusters (see \eqref{equ_def_cluster_partition}). 
	In contrast to Figure \ref{fig:ifca}, here IFCA is run with a miss-specified number 
	of clusters ($2$ clusters).\label{fig:ifca_no_clusters_not_known} }
\end{figure*}

\subsection{Comparison with Oracle Method}

This experiment compares the performance of Algorithm \ref{alg_pfl_regretmin} with an oracle approach that has perfect knowledge of the cluster structure \eqref{equ_def_cluster_partition}. 
Instead of the candidate selection rule in step \ref{equ_select_candidate_param} of 
Algorithm \ref{alg_pfl_regretmin}, the oracle approach randomly samples a node $\nodeidx' \in \cluster$ 
from the correct cluster $\cluster \ni \nodeidx$. 

For this experiment, we generated $\nrnodes = 100$ local datasets according 
to \eqref{equ_def_true_linear_model_SBM} with a local sample size $\localsamplesize{\nodeidx}\!=\!\samplesize\!=\!10$ 
and a local model dimension $\dimlocalmodel=2\!\cdot\!\samplesize$. 
The local datasets form $\nrcluster=2$ equal-sized clusters according to \eqref{equ_def_cluster_partition}. 
Figure \ref{fig:param_oracle} depicts the MSE of the oracle approach 
against Algorithm \ref{alg_pfl_regretmin}. We used a candidate set size of 
$\nrcandidates=20$. As shown in Figure \ref{fig:param_oracle}, Algorithm \ref{alg_pfl_regretmin} performs comparably to the oracle approach.

\begin{figure}[htbp]
	\begin{center}
       \ifshowtikz
		\begin{tikzpicture}[scale=0.8]
			\begin{axis}[
				xlabel={iteration nr.\ $\itercntr$},
				ylabel={MSE},
				ymajorgrids=true,
				grid style=dashed,
				table/col sep=comma,
				ymode=log,
				cycle list={
					cyan!40!blue!60,
					orange,
					black!70!green!70,
					purple,
					violet
				},
				]   
				\foreach \i/\j in {1/Algorithm \ref{alg_pfl_regretmin}, 2/ oracle }{
					\edef\temp{\noexpand\addlegendentry{\j}
						\noexpand\addplot+ [mark=none, very thick, solid] table [
						x index=0, 
						y index=\i] {Algo1_oracle.csv};
					}
					\temp}
			\end{axis}
		\end{tikzpicture}
	\end{center}
	\caption{
		Comparing the MSE incurred by Algorithm \ref{alg_pfl_regretmin} with the 
		MSE of an oracle approach.} 
	\label{fig:param_oracle}
	\vspace{-3mm}
\end{figure}

\subsection{Online Learning of Personalized Linear Model} 

We next study an online variant of Algorithm \ref{alg_pfl_regretmin} (see Section \ref{sec_online_variants}). 
This variant is particularly useful when the local datasets $\localdataset{\nodeidx'}$ are generated 
sequentially over time. It is then not possible to compute the gradient in 
\eqref{equ_def_gd_step} exactly (unless we wait until the very last data point). 
Instead, we use the most recent batch $\batch$ of data points to compute a 
gradient estimate $\mathbf{g}^{(\nodeidx')} \defeq  (-2/|\batch|) \sum_{\pair{\featurevec}{\truelabel} \in \batch}
 \featurevec\big(\truelabel - \featurevec^{T} \widehat{\weights} \big)$. 
 
We obtain an online variant of Algorithm \ref{alg_pfl_regretmin} by replacing the 
exact gradient $\nabla \locallossfunc{\nodeidx'}{\widehat{\weights}}$ in its step 
\ref{equ_def_gd_step} with the estimate $\mathbf{g}^{(\nodeidx')}$. Figure 
\ref{fig_loss_resample} illustrates the MSE incurred by the online variant of Algorithm \ref{alg_pfl_regretmin} 
with batch size $|\batch|\!=10$. As depicted in Figure \ref{fig_loss_resample}, 
this modification accelerates convergence for the cases when the ratio 
$\dimlocalmodel/\samplesize$ is not too small.

\begin{figure}[htbp]
	\begin{center}
		\begin{tikzpicture}[scale=0.8]
			\begin{axis}[
				legend style={nodes={scale=0.8}},
				xlabel={iteration nr.\ $\itercntr$},
				ylabel={MSE},
				ymajorgrids=true,
				grid style=dashed,
				table/col sep=comma,
				ymode=log,
				cycle list={
					cyan!40!blue!60,
					orange,
					black!70!green!70,
					purple,
					violet
				},
				]   
				\foreach \i/\j in {1/0.2, 2/1, 3/2, 4/5, 5/10}{
					\edef\temp{						\noexpand\addlegendentry{$\dimlocalmodel/\localsamplesize{\nodeidx}$ = \j}
						\noexpand\addplot+ [mark=none, very thick, solid] table [
						x index=0, 
						y index=\i] {Algo_1_resample.csv};
					}
					\temp}
			\end{axis}
		\end{tikzpicture}
	\end{center}
	\caption{
		MSE incurred by an online-variant of Algorithm \ref{alg_pfl_regretmin} with each iteration using gradient estimates computed from a fresh batch of $\samplesize$ data points.} 
	\label{fig_loss_resample}
\end{figure}

\subsection{Training a Personalized Decision Tree} 
\label{sec_exp_non_parametric}

This experiment examines the application of Algorithm \ref{alg_pfl_regretmin_modelagnostic} 
to train a personalized decision tree for client $\nodeidx=1$. The decision tree is constrained to a maximum depth $3$ and is trained using the synthetic local datasets \eqref{equ_def_true_linear_model_SBM} forming 
$\nrcluster=2$ clusters (see \eqref{equ_def_cluster_partition}). For the first client, with 
local dataset $\mathcal{D}^{(1)}$, we also generate a validation set $\mathcal{D}_{\rm val}^{(1)}$ 
of size $m_{\rm val} = 100$. This validation set is used to evaluate the decision tree delivered by Algorithm \ref{alg_pfl_regretmin_modelagnostic} \cite{JMLR:v12:pedregosa11a}.
 
The initialization $\learnthypothesis$ for Algorithm \ref{alg_pfl_regretmin_modelagnostic} 
is obtained by training a decision tree locally, using only $\mathcal{D}^{(1)}$. 
The regularization parameter value is set to $\eta=1$, and the test set $\testset$ comprises $m_{t}=100$ i.i.d. realizations $\vx^{(1)},\ldots,\vx^{(100)} \sim \mathcal{N}(\mathbf{0},\mathbf{I})$
of a standard multivariate normal distribution.

We implement the update \ref{equ_def_update_modelagnostic}, for a candidate $\nodeidx'$, 
by training a decision tree on the augmented dataset 
\begin{equation}
\big\{ \pair{\featurevec}{\widehat{\hypothesis}(\featurevec)} \big\}_{\featurevec \in \testset}	 \cup  \localdataset{\nodeidx'}. 
\end{equation} 
The training was implemented using the \texttt{DecisionTreeRegressor.fit()} function of 
\texttt{scikit-learn} \cite{JMLR:v12:pedregosa11a}. This function allows to 
specify individual weights for each data point in the training set. 

We also train oracle model on all $50$ local datasets belonging to the same cluster as the first 
client's dataset $\mathcal{D}^{(1)}$. To measure the quality of the learnt local model 

$\widehat{\hypothesis}$ we use normalized mean squared error
\begin{equation}
		\label{equ_def_MSE_nonparam_norm}
{\rm MSE_{norm}} \defeq 	\frac{\rm MSE}{\rm MSE_{oracle}}.
\end{equation}
where $\rm MSE$ is defined as
\begin{equation}
		\label{equ_def_MSE_nonparam}
{\rm MSE} \defeq 	\sum_{\pair{\vx}{y} \in \mathcal{D}_{\rm val}^{(1)} } \normgeneric{\widehat h(\mathbf{x}) -  y }{2}^{2}.
\end{equation}
and 
\begin{equation}
	\mathcal{D}_{\rm val}^{(1)} = \bigg\{ \left({\mathbf{x}^{(1)}}, y^{(1)} \right), ..., \left({\mathbf{x}^{(m_{\rm val})}}, y^{(m_{\rm val})} \right)  \bigg\}
\end{equation}
and $\rm MSE_{\rm oracle}$ denotes the MSE of the oracle model incurred on the validation set $\mathcal{D}_{\rm val}^{(1)}$.

We apply Algorithm \ref{alg_pfl_regretmin_modelagnostic} to different instances of the 
local linear model \eqref{equ_def_true_linear_model_SBM} obtained for $\dimlocalmodel \in \{2,10,20,50,100 \}$ 
and present the results in Figure \ref{fig:nonparam}. Alongside the ${\rm MSE_{norm}}$ defined above (solid lines in the plot), normalized MSE values for models trained locally — only using the training set $\mathcal{D}_{\rm train}^{(1)}$ — are also plotted. These normalized 
MSE values are depicted with dashed lines in Figure \ref{fig:nonparam}, with colours corresponding to specific $\dimlocalmodel/\samplesize$ ratios. As shown in Figure \ref{fig:nonparam}, Algorithm \ref{alg_pfl_regretmin_modelagnostic} performs comparably or slightly worse than the oracle model but generally better than or similar to the locally trained model.

\begin{figure}[htbp]
	\begin{center}
		\begin{tikzpicture}[scale=0.8]
			\begin{axis}[
                legend style={nodes={scale=0.8}},
				xlabel={iteration nr.\ $\itercntr$},
				ylabel={$\rm MSE_{norm}$},
				ymajorgrids=true,
				grid style=dashed,
				table/col sep=comma,
				ymode=log,
				cycle list={
					cyan!40!blue!60,
					orange,
					black!70!green!70,
					purple,
					violet
				}
				]   
				\foreach \i/\j in {1/0.2, 2/1, 3/2, 4/5, 5/10}{
					\edef\temp{
						\noexpand\addlegendentry{$\dimlocalmodel/\localsamplesize{\nodeidx}$ = \j}
						\noexpand\addplot+ [mark=none, very thick, solid] table [
						x index=0, 
						y index=\i] {Algo_2_mse.csv};
					}
					\temp}
                \foreach \i in {1, 2, 3, 4, 5}{
    					\edef\temp{
    						\noexpand\addplot+ [mark=none, very thick, dashed] table [x index=0,
    						y index=\i] {Algo_2_mse_baseline.csv};
    					}
    					\temp}
				\addplot[black, very thick, dashed, domain = -10:500, samples = 2] {1};
				
			\end{axis}
		\end{tikzpicture}
	\end{center}
	\caption{Normalized MSE incurred by the decision tree learnt by Algorithm \ref{alg_pfl_regretmin_modelagnostic} 
		from the local datasets \eqref{equ_def_true_linear_model_SBM} for varying $\dimlocalmodel/\localsamplesize{\nodeidx}$. 
		Dashed lines correspond to performance of the model trained locally.} 
	\label{fig:nonparam}
\end{figure}

\section{Conclusion}

We studied an active sampling approach for distributed learning of personalized models in a 
clustered local datasets setting. PersFL is a scalable FL method based on the assumption 
that the gradient, or its generalization as described in \ref{equ_def_update_modelagnostic}, 
conveys information about the similarities between the local datasets. 

PersFL learns personalized models by sampling useful candidate datasets from the 
collection of local datasets and uses these datasets to incrementally improve the model at hand.

This approach is particularly effective when local datasets are heterogeneous, as centralized FL 
methods that average model parameters or gradients often result in deteriorated performance in 
such scenarios  \cite{pmlr-v195-zhao23b}. Furthermore, our method does not require the knowledge 
of the similarity graph of local datasets or the number of clusters they belong to, and it can be 
extended to non-parametric cases.   

\section{Acknowledgements}

This work has been supported by Research Council of Finland under decision nrs. 331197, 349966, 363624.

\bibliography{sn-bibliography}

\end{document}